# Intelligent City Traffic Management and Public Transportation System

**S.A.Mulay[1], C.S.Dhekne[2], R. M. Bapat[3], T. U. Budukh[4], S. D. Gadgil[5]**

[1]**Professor: Department of Computer, PVG's COET**
**Pune, Maharashtra, India**

[2,3,4,5] **Students: Department of Computer, PVG's COET**
**Pune, Maharashtra, India**

## Abstract

Intelligent Transportation System in case of cities is controlling traffic congestion and regulating the traffic flow. This paper presents three modules that will help in managing city traffic issues and ultimately gives advanced development in transportation system. First module- Congestion Detection and Management will provide user real time information about congestion on the road towards his destination, second module – Intelligent Public Transport System will provide user real time public transport information(local buses), and the third module- Signal Synchronization will help in controlling congestion at signals, with real time adjustments of signal timers according to the congestion. All the information that user is getting about the traffic or public transportation will be provided on user's day to .day device that is mobile (through Android application/SMS). Moreover, communication can also be done via Website for Clients having internet access. And all these modules will be fully automated without any human intervention at server-side

**Keywords**: *Intelligent Transportation Systems, Intelligent Public Transport System, Traffic Signal Synchronization Techniques.*

## 1. Introduction

This paper mainly focuses on the concept of Intelligent Transportation System (ITS), covering domains like evolutionary computing and intelligent systems, mobile computing and applications, image processing, internet service and applications, GPS and location based applications, web technologies, internet services and applications Generally during peak hours, there are a huge number of vehicles running on the road and we find very high density of vehicles at traffic junctions. Hence everyone has to wait for many turns to pass through that junction. Generally at office or School timings people are in hurry to reach their workplaces on the time. And if they have to wait for long times at signals, they start doing invalid moves like breaking signals , breaking traffic rules etc. and thus traffic situations become more and more critical to handle. The goal of ITS is to improve the transportation system to make it more effective, efficient and safe.

### 1.1 Aims

The aim is to develop Datacenter for city ITS. City ITS is nothing but Intelligent Transportation System related to a particular city.

Here we can make transportation in city very effective and efficient with the help of ITS concept. Important need in city is to control traffic in congestion areas. We often see traffic jams at peak timings and it takes a very long time to resolve itself. Hence to regulate traffic effectively in very short time is the necessary real time task.

### 1.2 Overview

- *Intelligent Public Transport Automation:*

User is notified about the real-time position of the bus and this information is provided on user's mobile phone. Real-time position of the bus is tracked by GPS Tracker located in the bus.

- *CCTV Image Processing*:

CCTV camera is located at every chowk. Capture the CCTV Images and store them in database in appropriate format [1]. So, according to these images, we are calculating congestion results at that particular road and providing them to users. This information is provided on user's mobile phone. Virtual map of that road with proper color coding is given to the user.

## 2. Explanation of methods

### 2.1. Intelligent public transport system

- *Working at Data-Center:*

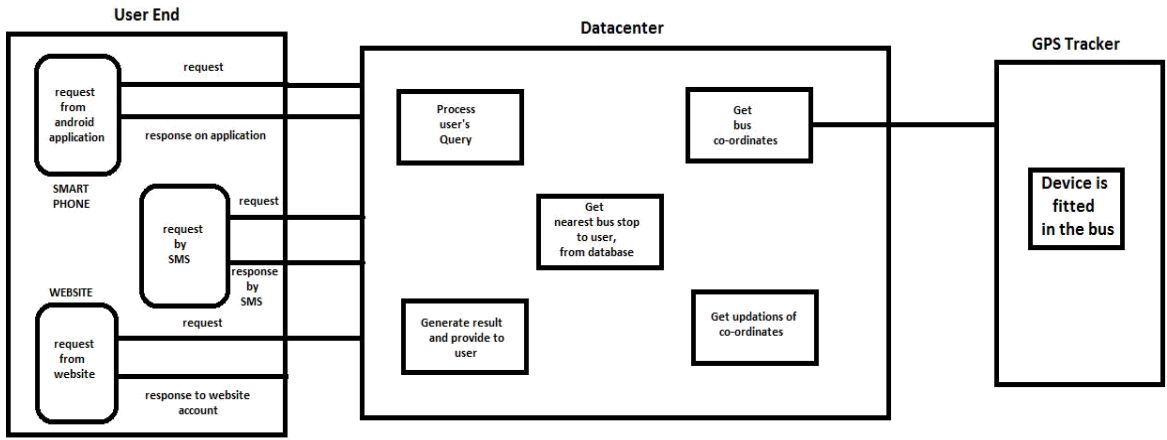

Fig. 1 Working Flow for Public Transport Automation

Data-center will constantly track down real-time position of all buses and it will update those positions in the database in a pre-determined interval of timer. As user-query arrives at data-center, according to the source and destination of the user, relative bus information such as bus-no proceeding from user's source to destination, distance of that bus from user's nearest bus-stop to his source, approximate arrival time of the bus to the nearest bus-stop is provided back to the user. Refer Fig.1 [10]

- *User's Request from Website :*

Firstly user creates his account with unique username and password. After this he is getting registered to the service. Now user can query with filling source and destination to the service. And he is getting bus information towards his destination.

- *User's Request from Android Application:*

Similar to website, when user logins into the application, he just he will simply enter the desired destination to get bus information towards that destination. His source will be automatically traced by GPS tracker placed in his smart phone. If GPS service is OFF, then user's network location will be considered as his source. Refer Fig.2 and Fig.3

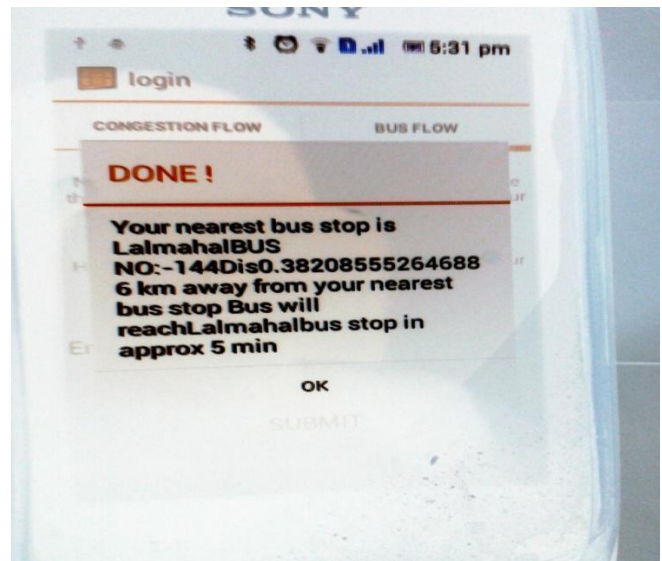

Fig.3 Response to Android Application

- *User's Request from SMS:*

Here in this flow user has to send message (SMS) to a particular service number with some typical query format. This query format includes source of the user and user has to provide his destination, where he wants to go. Refer fig.4 and fig.5. 'BUS' represents query for congestion and source of user is 'AB Chowk' and destination is 'Nal Stop'. ( Sim card having number +919766429259 is inserted in GSM modem, hence user will request to that number and he will receive response from the same number)

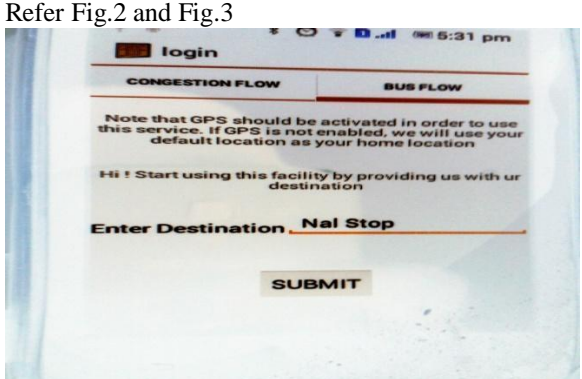

Fig.2 User query from Android Application

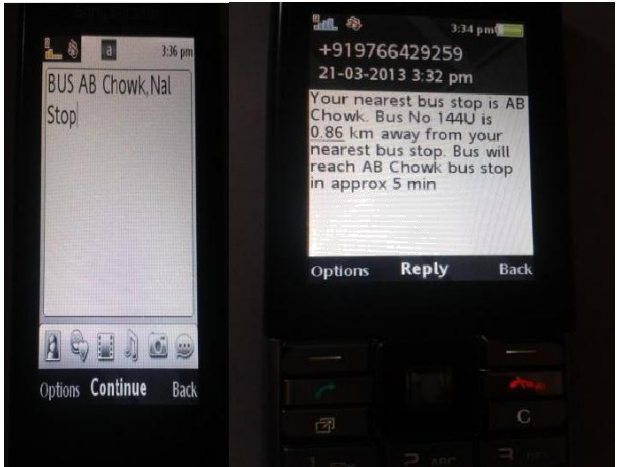

Fig.4 SMS query   Fig.5 SMS response

## 2.2 Traffic congestion detection using CCTV image processing

As we all know, there is large amount of vehicle number, increasing day by day in the city. Especially during festive seasons, traffic conditions are so bad and they are just very much difficult to control and traffic jams happen frequently. Hence it is necessary to control congestion in getting help to simplify traffic and make the traffic flow smooth. Basic idea to control traffic can be diverting traffic. Hence increasing amount of congestion is the main need of proposal of this idea. Refer Fig.9 [10]

- *Image Processing At Datacentre:*

Images from CCTV cameras are captured and store them in database in appropriate format .These images are captured after a specific time interval. In order to find out traffic density of any road, we follow the following steps:

Firstly we capture the image of empty road. This image is used for calculating total width of that road excluding parking space. We suppose that the camera is at the *chowk*. Successive images will be captured after a specific time interval and the latest captured image will be compared to empty road image and the previously captured image to get relative amount of congestion.

Image comparison will be done in following way:

Empty road image and latest captured image are getting compared pixel by pixel. Pixels which are same, getting filled as of color white in another image called processed image and pixels which are different, getting filled as a color of red in that processed image. Hence we are now getting third image that is processed image, which is a difference of that the two images comparison. Refer Fig. 6, 7,8(model for demonstration images). Also we are comparing two successive images to get relative difference in traffic. This comparison of successive images is also done with the same technique.

Artificial intelligence is also introduced in this idea. As we all know that congestion amount on Sunday is different from congestion amount on Monday. Hence there must be some trend of each day of week. Hence traffic situations on every day of week and at every hour of that day are getting captured from CCTV camera. This information is getting stored in our database. Datacentre is using this data and historical data also to get traffic trend of a particular day of a week. Hence we can also provide in advanced approximate value of congestion on particular day of week and particular time of that day.

In this way, Datacenter is receiving static images from CCTV cameras placed at different locations. Congestion statistics and relative traffic congestion percentage are getting calculated at some typical time interval and this real time information is provided to user as per user's request. Datacentre is continuously processing images and calculating congestion results even if there is no congestion request from user end. And if user sends wrong or inappropriate query, he is informed with error message and appropriate query format.

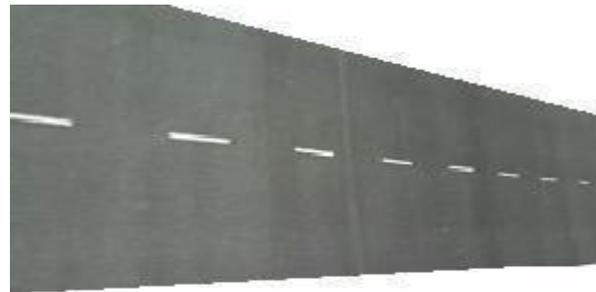

Fig. 6 Empty road image

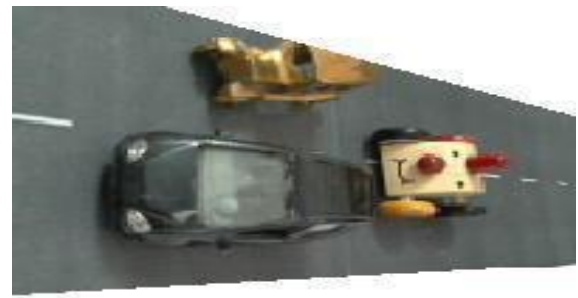

Fig.7 Image with Vehicles

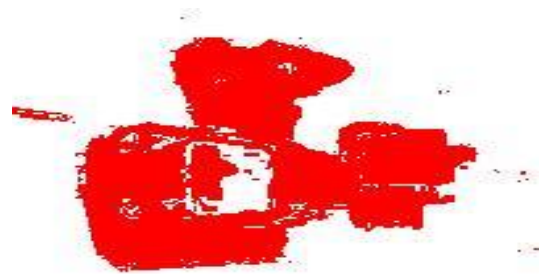

Fig.8 Processed image showing difference of comparison

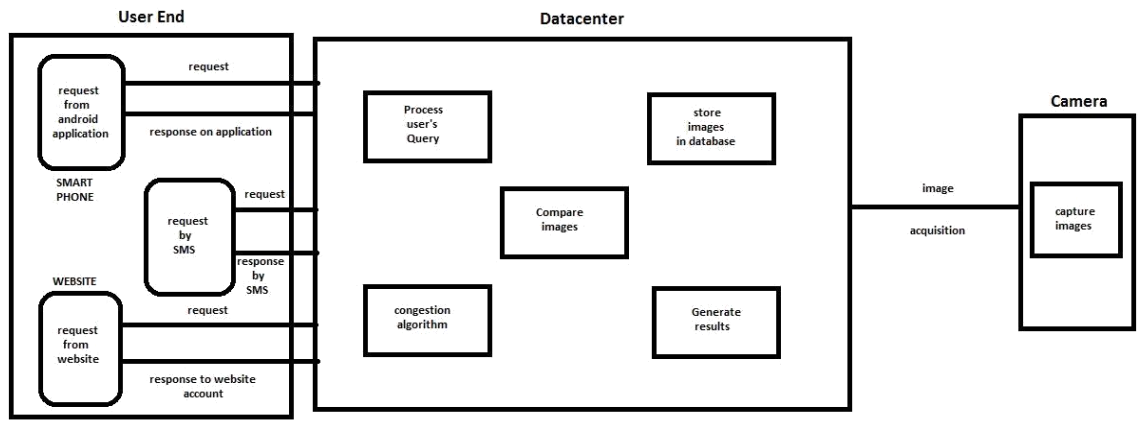

**Block Diagram for congestion detection**

Fig 9 Working Flow of Congestion Detection.

- *User's Request from Website :*

Firstly user creates his account with unique username and password. After this he is getting registered to the service. He can now make query by entering his source and destination. He can also get virtual map with proper color coding according to congestion. He needs to click view virtual map button. Virtual map is the map which depicts a particular road.

- *User's Request from Android Application :*

This is same as Android Application request for bus information.
Query format: Enter destination.
Response format: Exact congestion Percentage, trend of last 5 minutes, traffic status.

- *User's Request from SMS:*

This is same as SMS request for bus flow.
Query format: Provide source and destination
Response format: Exact congestion Percentage, trend of last 5 minutes, traffic status.

2.3 Signal Synchronization

In this module, Accurate green and red timings for signal are devised according to the congestion present at that road at that particular time. This will help in resolving congestion effectively. This change in timings is not abrupt. There is gradual change and the traffic is resolved efficiently.

If the congestion is detected at more than one road at a time, in the same *chowk* the percentage of congestion will get compared at data centre and road having comparatively greater congestion will get the priority. Not only the congestion but also the factors like length, width of the road, daily trend of that road is considered.

## 3. Data Used

3.1 CCTV Images

Images captured from CCTV cameras are processed effectively, to get congestion results at particular road.

3.2 Database for Bus

We can use and store data of every bus with its bus number, route and arrival times at every stop to the database. This information is collected from the web and from the news paper displaying complete data of 'Pune Bus Day Program'.

3.3 GPS Information

Real time GPS co-ordinates of bus are getting constantly updated in database. Using this data of co-ordinates we are calculating results of the particular bus that user requested for.

## 4. Instrumentation

4.1 CCTV Camera

In this paper CCTV camera is basically used for inputting data to the Datacenter in the form of images. These cameras are wired, as wired cameras give better accuracy than wireless cameras. These cameras are placed at *chowk* for a road with its range of vision.

4.2 GSM Modem

GSM modem is used to receive user's query about congestion, bus via SMS. GSM modem is very much effective to make communication between Datacenter and users those are not having smart phones.

4.3 GPS Tracker

GPS tracker is used to get the current bus co-ordinates. Once we get these co-ordinates of the bus we can do further calculations about arrival time of bus considering congestion. The GPS tracker will inform current and real time positions of the bus. This device is placed in the bus only.

## 5. Results

Following are the results of implementation of three modules:

5.1 Intelligent Public Transport System
1. User is getting real time information of the bus travelling towards his desired destination.
2. User is getting bus number, nearest bus stop to his source and the arrival time of the bus to that bus stop.

So just by providing his source and destination, he is able to view public transport information on his mobile.

5.2 Congestion Detection and Management:
1. User is getting real time information of congestion on the road towards his destination as exact congestion percentage on the road, trend of last 5 minutes and traffic status, by requesting through website/Android Application/SMS.
2. User is getting virtual map on web site and on Android Application. So that he can get graphical scenario of that road.

5.3 Signal Synchronization:
1. Signal timers are adjusted according to congestion detected at every traffic junction.

Real time signal timing adjustment is done to avoid traffic jams at signals.

## 6. Conclusion

This paper focuses on designing and developing datacentre which is working without any human intervention. This datacentre is automatic and autonomous. This automatically eliminates human errors and decreases the risk of application malfunction. As CCTV cameras are being used to detect traffic congestion, there are no blind spots. It is extremely useful to users as he is getting all the information on his daily-used device that is mobile phone. All traffic updates will be obtained based upon the real-time analyzed data hence user can take a decision whether to travel by that route or to opt for some alternative route. Real time bus information will help user in reaching bus stop on-time. And his unnecessary wait at bus stop is avoided. This will improve the efficiency of Bus Transport. Ultimately the pollution and fuel consumption will be controlled. Signal synchronization will indirectly help the user as only short time wait at signals and will help in resolving high traffic density, especially at rush timings. In this way datacentre will help in controlling congestion, facilitating user with traffic status and bus information.

And city transportation system becomes intelligent and efficient.

## 8. Biography

**S.A.Mulay** : B.E. C.S.E.,2001, M.Tech. I.T.,2011.
Lecturer, D.K.T.E.'s Textile & Engg. Institute, Ichalkaranji, Maharashtra, India.
Assi.Prof.,PVG's College of Engg., Pune, Maharashtra, India. Cambridge International Certification for Teachers & trainers. No. of papers published: 3.Research Interest : Cloud Computing

**C.S.Dhekne[2], R. M. Bapat[3], T. U. Budukh[4], S. D. Gadgil[5]**: Pursuing B.E Computer Engineering at PVG's COET, Pune. Best Paper Award at ICACSIT, Chennai for paper titled "Concept Based Implementation of a Datacenter for Intelligent City Traffic Management".